\documentclass[11pt]{article}
\usepackage{acl2002}
\usepackage{amsmath}
\usepackage{amssymb, latexsym}
\usepackage{gb4e}
\usepackage{algorithm}
\usepackage{algorithmic}
\usepackage{times, mathptm}

\allowdisplaybreaks[2]

\title{Unsupervised Learning of Morphology Without Morphemes}

\author{Sylvain Neuvel\\ Dept.\ of Linguistics\\ sneuvel@uchicago.edu \And 
Sean A. Fulop\\ Depts. of Linguistics\\ and Computer Science\\
sfulop@uchicago.edu \vspace{-10pt} \AND  {\rm The University of
Chicago}}

\begin{document}
\maketitle
\renewcommand{\textfraction}{0}

\begin{abstract}
The first morphological learner based upon the theory of
Whole Word Morphology \cite{Fordetal1997} is outlined, and preliminary evaluation results
are presented.
The program, Whole Word Morphologizer, takes a POS-tagged lexicon as
input, induces morphological relationships without attempting to
discover or identify morphemes, and is then able to generate new words beyond the learning
sample.
The accuracy (precision) of the generated new words is as high as 80\%
using the pure Whole Word theory, and 92\% after a post-hoc adjustment
is added to the routine.
\end{abstract}

The aim of this project is to develop a computational model employing
the theory of \emph{whole word morphology} \cite{Fordetal1997} capable on the one hand of identifying
morphological relations within a list of words from any one of a wide variety of
languages and, on the other, of putting that knowledge to use in
creating previously unseen word forms.
A small application called Whole Word Morphologizer which does just
this is outlined and discussed.
In particular, this approach is set against the literature on
computational morphology as an entirely different way of doing things
which has the potential to be generalized to all known varieties of
morphology in the world's languages, a feature not shared by previous
methods.
As it is based on a model of
the mental lexicon in which all entries are entire, fully
fledged words, this project also serves as an empirical demonstration
that a word-based morphological theory that rejects the notion of morpheme as
minimal unit of form and meaning (and/or grammatical properties) is
viable from the point of view of acquisition as well as generation.

\section{Morphological learning}

Since its inception in the mid 1950s, the field of computational
morphology has been characterized by a paucity of procedures for
generation.  Notwithstanding the impressive body of literature on the
shortcomings of traditional Paninian morphology, most computational
research projects also rely on a traditional notion of the morpheme
and ignore all non-compositional aspects of morphology. These
observations are obviously not unrelated and are in part inherited from
the field of computational syntax where applications traditionally were designed to assign a syntactic structure
to a given string of words, though this is less true today.

\subsection{Segmentation and morpheme identification}

Word formation and the population of the lexicon, while central to
morphological theory, are noticeably absent from the field of
computational morphology.  Most computational work in the field of
morphology has focused on the identification of morphemes or
morphological parsing while paying little or no attention to
generation.  While these applications find a common goal in the
automatic acquisition of morphology, it is helpful to
distinguish between two types of analysis in light of the often very
different results sought by various morphological learners.  

On the one hand, some applications focus exclusively on the
\textbf{segmentation} of words or longer strings into smaller units.  In other
words, their function is to identify morpheme boundaries within words
and, as such, they only indirectly identify morphemes as linguistic
units.  
Zellig Harris's \cite{Harris1955,Harris1967} pioneering work
suggests that morpheme boundaries can be determined by counting the
number of letters that follow a given substring within a corpus
(v.\ \cite{HaferWeiss1974} for a further development of Harris's ideas).
Janssen \shortcite{Janssen1992} and Flenner \shortcite{Flenner1994,Flenner1995} also work towards segmenting
words but use training corpora in which morpheme boundaries have been
manually inserted.  Recent work by Kazakov and Manandhar \shortcite{KazakovManandhar1998}
combines unsupervised and supervised learning techniques to generate a
set of segmentation rules that can further be applied to previously
unseen words.  

On the other hand, some computational morphological applications are
designed solely to \textbf{identify morphemes} based on a training corpus and
not to provide a morphological analysis for each word of that corpus.
Brent \shortcite{Brent1993}, for example, aims at finding the right
set of suffixes from a corpus, but the
algorithm cannot double as a morphological parser.

More recently, efforts have been developing which identify morphemes
\emph{and} perform some sort of analysis.
Schone and Jurafsky \shortcite{SchoneJurafsky2001} employ a great many
sophisticated post-hoc adjustments to obtain the right conflation sets
for words by pure corpus analysis without annotations.
Their procedure uses a morpheme-based model, provides an analysis of
the words, and does in a sense
discover morphological relations.
 Goldsmith \shortcite{Goldsmith2001a,Goldsmith2001b}, inspired by de
Marcken's \shortcite{deMarcken1995} thesis on minimum
description length,  attempts
to provide both a list of morphemes and an analysis of each word in a corpus.
Also, Baroni \shortcite{Baroni2000} aims at finding a set of prefixes
from a corpus, together with an affix-stem parse of each of the words.

While they might differ in their methods or objectives, all of the
above morphological applications share a common
characteristic in that they are learners designed exclusively
for the acquisition of morphological \emph{facts} from corpora and do not
generate new words based on the information they acquire.

\subsection{Parsing and generation}

Only a handful of programs can both parse and generate words.  Once
again, these programs fall into two very distinct categories.  In view
of the disparity between these programs, it is useful to distinguish
between genuine morphological learners able to generate from acquired
knowledge and generators/parsers that implement a man-made analysis.
The latter group is perhaps the most well known, so let us begin with them.

Kimmo-type applications of two-level morphology \cite{Koskenniemi1983,Antworth1990,Karttunenetal1992,Karttunen1993,Karttunen1994} can provide a morphological analysis of the
words in a corpus and generate new words based on a set of rules; but
these programs must first be provided with that set of rules and a
lexicon containing morphemes by the user.
  Similar work in one- and
two-level morphology has been done using the Attribute-Logic Engine
\cite{Carpenter1992}.
Some of these systems (e.g.\ \cite{Karttunenetal1987}) have a front-end that compiles more
traditional linearly ordered morphological rules into the finite-state automata of
two-level morphology.
  Once again, these
applications require a set of man-made lexical rules to function.
While the practical uses of such applications as PC-Kimmo are
incontestable, it is clear that they are part of a different
endeavour, and should not be confused with genuine morphological
learners.  

The other relevant group of computational applications can, as
mentioned, both acquire morphological knowledge from corpora and
generate new words based on that knowledge.  Albright and Hayes
\shortcite{AlbrightHayes2001a,AlbrightHayes2001b} tackle the wider task of acquiring morphology
and (morpho)phonology based on a small paradigm list and their learner
is able to generate particular inflected forms given a related word.
D\v{z}eroski
and Erjavec \shortcite{DzeroskiErjavec1997} work towards learning morphological rules for
forming particular inflectional forms given a lemma (a set of related words).  
Their learner
produces a set of rules relating all the members of a paradigm to a
base form.  The program can then produce a member of that paradigm on
command given the base form.  While the methods used by Albright and
Hayes and D\v{z}eroski and Erjavec radically differ, both use a form of
supervised learning which significantly reduces the amount
of information their learner has to acquire.
  Albright and Hayes train
their program using a paradigm list in which each entry contains, for
example, both the present and past tense forms of an English verb.
Similarly, the training data used by D\v{z}eroski and Erjavec similarly
has a base form, or lexeme, associated to each and every word so that
all the words of a given paradigm share a common label.  
The distinctions between the two methods are immaterial, what matters is
that both learners are being told which words are related to which
and are left with the task of describing that relation in the
form a rule. 
 In other words, the algorithms they use cannot discover
that words are morphologically related. 

\subsection{What's morphology?}

In the above algorithms, the task of determining
whether one word is related to another in a morphological sense is
most frequently left to the linguist, as this information has to be
encoded in the training data for these algorithms.
(Some of the most recent work such as \cite{SchoneJurafsky2001} and
\cite{Goldsmith2001a} are notable exceptions to this paradigm.)
  This is perhaps not surprising, since
no serious attempt at defining a morphological relation has been made
in the last few decades.  American structuralists of the forties and
fifties proposed what have been referred to as discovery procedures
(v.\ \cite{Nida1949}, for example) for the identification of morphemes but
since the mid fifties \cite{Chomsky1955}, it has been customary for
morphological theory to ignore this aspect of morphology and relegate
it to studies on language acquisition.  
But, since a morphological learner like that presented here is
designed to model the acquisition of morphology, it seems that it
should above all be able to determine \emph{for itself}
whether two words are morphologically related or not, whether there is
anything morphological to acquire at all.  

Another important thing to note about the vast majority of computational
morphology learners is their reliance on a traditional notion
of the morpheme as a lexical unit and their exclusive focus on
concatenative morphology.  
There is a panoply of recent publications
devoted to the empirical shortcomings of traditional so-called
``Item-and-Arrangement'' morphology \cite{Hockett1954,Bochner1993,FordSingh1991,Anderson1992,Fordetal1997}, and the list of
phenomena that fall out of reach of a compositional approach is rather
impressive:  zero-morphs, ablaut-like processes, templatic morphology,
class markers, partial suppletion, etc.
Still, seemingly every documented morphological learner relies on a Bloomfieldian
notion of the morpheme and produces an Item-and-Arrangement analysis;
this description applies to all of the computational papers cited
above.

\section{An alternative theory}

Whole Word Morphologizer (henceforth WWM) is the first implementation
of the theory of Whole Word Morphology.  The theory, developed by Alan
Ford and Rajendra Singh at Universit\'{e} de Montr\'{e}al, seeks to account
for morphological relations in a minimalist fashion.
Ford and Singh published a series of papers
dealing with various aspects of the theory between 1983 and
1990. Drawing on these papers, they published a full outline of it in
1991 \cite{FordSingh1991} and an even fuller defense of it in 1997
\cite{Fordetal1997}.
 Since then, aspects of
it have been taken up in a series of publications by Agnihotri,
Dasgupta, Ford, Neuvel, Singh, and various combinations of these
authors.  The central mechanism of the theory, the Word Formation
Strategy (WFS), is a sort of non-decomposable morphological
transformation that relates full words with full words (or helps one
fashion a full word from another full word) and parses
any complex word into a variable and a non-variable component.  Neuvel
and Singh \shortcite{NeuvelSingh2002} offer a strict definition of morphological relatedness
and, based on this definition, suggest guidelines for the acquisition
of Word Formation Strategies.

In Whole-Word Morphology, any morphological
relation can be represented by a rule of the following form:
\begin{exe} 
\ex \label{relation} $\lvert X \rvert_\alpha \leftrightarrow \lvert X' \rvert_\beta$ 
\end{exe}
in which the following conditions and notations are employed:
\begin{enumerate}
\item $\lvert X \rvert_\alpha$ and $\lvert X' \rvert_\beta$ are statements that words of the form $X$ and $X'$ are possible in the language,
and $X$ and $X'$ are abbreviations of the forms of classes of words
belonging to categories $\alpha$ and $\beta$ (with which specific
words belonging to the right category can be unified in form);
\item $'$ represents all the form-related differences between $X$ and $X'$;
\item $\alpha$ and $\beta$ are categories that may be represented as feature-bundles;
\item $\leftrightarrow$ represents a bi-directional implication;
\item $X'$ and $X$ are semantically related.
\end{enumerate}

There are several ramifications of (\ref{relation}). 
First, there is only
one morphology; no distinction, other than a functional one, is made
between inflection and derivation.
Second, morphology is relational and
not compositional. 
The program thus
makes no reference to theoretical constructs such as `root', `stem',
and `morpheme', or devices such as `levels' and `strata' and relies
exclusively on the notion of morphological relatedness.
 And since its
objective is not to assign a probability to a given word or string, it
must rely on a strict formal definition of a morphological
relation. 
Ultimately, the theory takes the Saussurean view
that words are defined by the differences amongst them and argues that
some of these differences, namely those that are found between two or
more pairs of words, constitute the domain of morphology.  In other
words, two words of a lexicon are morphologically related if and only
if all the differences between them are found in at least one other
pair of words of the same lexicon.

\section{Overview of the method}

Under the assumption that the morphology of a language resides
exclusively in differences that are exploited in more than one pair of
words within its lexicon, WWM (Algorithm~\ref{WWMalg} in the next section) compares every word of a small lexicon
and determines the segmental differences found between them.  The
input to the current version of the program is a small text file that
contains anywhere from 1000 to 5000 words. Each word appears in
orthographic form and is followed by its syntactic and morphological
categories, as in the example below:

\begin{exe}
\ex \begin{tabular}[t]{lll}
cat,&Ns&(Noun, singular)\\
catch,&V&\\
catches,&V3s&\parbox[t]{1.5in}{(Verb, (pres.) 3rd pers.\ sing.)}\\
decided,&Vp&(Verb, past)
\end{tabular}
\end{exe}

The algorithm simply compares each letter from word A to the
corresponding one from word B to produce a comparison record, which
can be viewed as a data structure.
Currently, it works on orthographic representations.
This means it would as easily work on phonemic transcriptions, but it
will require empirical evaluation to see whether the results from
these can improve upon those obtained using spellings, and we have not yet
gone through such an exercise.
It starts on either the left or right edge of the words if the two words share
their first (few) segments or their last (few) segments, respectively
(the forward version is presented in Algorithm~\ref{compforward} in
the next section).
This is just a simple-minded way of aligning the similar parts of the
words for the comparison; a more sophisticated implementation in the
future could use a more general sequence alignment procedure.
The segments are placed in one of two lists in the comparison structure (differences or similarities) based
on whether or not they are identical. Each comparison structure also contains
the categories of both words, and is kept in a large list of all
comparison structures found from analyzing the entire corpus.
The example below shows the information in the comparison structure
produced from the English words \emph{receive} and \emph{reception.}
It includes the differences and similarities between the two words,
from the perspective of each word in turn, as well as the lexical
categories of the words.
\begin{exe}
\ex \begin{tabular}[t]{cc}
\multicolumn{2}{c}{Differences}\\
First word&Second word\\[5pt]
\#\#\#\#ive$_{\rm V}$&\#\#\#\#ption$_{\rm Ns}$\\[8pt] \hline 
\multicolumn{2}{c}{Similarities}\\
First&Second\\[5pt]
rece\#\#\#&rece\#\#\#\#\#
\end{tabular}
\end{exe}

Matching character sequences in the difference section are replaced with a
variable. The result is then set against comparisons generated by other pairs of words and duplicate
differences are recognized. In the example below, the comparisons
produced by the pairs \emph{receive/reception, conceive/conception} and
\emph{deceive/deception} are shown.

\begin{exe}
\ex \label{eption} 
\begin{tabular}[t]{cc}
\multicolumn{2}{c}{Differences}\\
First word&Second word\\[5pt]
X ive$_{\rm V}$&X ption$_{\rm Ns}$\\
X ive$_{\rm V}$&X ption$_{\rm Ns}$\\
X ive$_{\rm V}$&X ption$_{\rm Ns}$\\[8pt] \hline
\multicolumn{2}{c}{Similarities}\\
First&Second\\[5pt]
rece\#\#\#&rece\#\#\#\#\#\\
conce\#\#\#&conce\#\#\#\#\#\\
dece\#\#\#&dece\#\#\#\#\#
\end{tabular}
\end{exe}

The three comparisons in (\ref{eption}) share the same \emph{formal and
grammatical} differences, and so the theory indicates they should be merged into one morphological
strategy.
Since the differences are the same, it is only the similarities that
are actually merged.
 Each new morphological strategy is also restricted to apply
in as narrow an environment as possible. Neuvel and Singh \cite{NeuvelSingh2002}
suggest that any morphological strategy must be maximally restricted
at all times; this is accomplished by specifying as constant all the
similarities found, not between words, but between the similarities
found between words.
 In (\ref{eption}), all three sets of similarities end with
the sequence of letters ``ce.''
 These similarities
between similarities are specified as constant in each strategy and
the length of each word is also factored in.
The \texttt{merge} routine called in Algorithm~\ref{compforward}
carries out this procedure; we don't show it because it is tedious but
not especially interesting.
The restricted
morphological strategy relating the words in (\ref{eption}) is as follows: 
\begin{exe}
\ex \label{eption2}
\begin{tabular}[t]{cc}
\multicolumn{2}{c}{Differences}\\
First word&Second word\\[5pt]
X ive$_{\rm V}$&X ption$_{\rm Ns}$\\[8pt] \hline
\multicolumn{2}{c}{Similarities}\\
First&Second\\[5pt]
$*$\#\#ce\#\#\#&$*$\#\#ce\#\#\#\#\#
\end{tabular}
\end{exe}

For the sake of clarity, we can represent the information contained in
 (\ref{eption2}) in a more familiar fashion using the formalism
 described in (\ref{relation}).
 The vertical brackets `$\lvert \cdot \rvert$' are used for orthographic
 forms so as not to confuse them with phonemic representations.
\begin{exe}
\ex \label{eption3}
 $\lvert\textrm{$*$\#\#ceive}\rvert_{\rm V} \leftrightarrow \lvert\textrm{$*$\#\#ception}\rvert_{\rm Ns}$
\end{exe}

The `\#' signs in the above representations stand for letters that must
 be instantiated but are not specified; the `$*$' symbol stands for a
 letter that is not specified and that may or may not be instantiated.
 Strategy (\ref{eption3}) can therefore be interpreted as follows: 
\begin{exe}
\exp{eption3} If there is a verb that ends with the sequence ``ceive'' preceded by no less than two and no more than three characters, there should also be a singular noun that ends with the sequence ``ception'' preceded by the same two or three characters.
\end{exe}
After performing the comparisons and merging, WWM extracts a list of
morphological strategies, which are those comparison structures whose
count is more
than some fixed threshold.
Table~\ref{Moby} contains a few strategies found from the first few chapters of \emph{Moby Dick.}
These strategies result from merging comparison structures which have
the same differences---merging the similarities of several
unifiable word pairs, and so many have no specified letters at all.
\begin{table*} \caption{Word-formation strategies discovered from \emph{Moby
Dick}}
\label{Moby}
\medskip
\begin{tabular}{|ll|ll|l|}
\hline \multicolumn{2}{|c|}{Differences}&\multicolumn{2}{|c|}{Similarities}&\\
1st word&2nd word&1st word&2nd word&Examples\\ \hline
Xd$_{\rm PP}$&X$_{\rm V}$&$*$$*$$*$$*$\#\#\#\#e\#&$*$$*$$*$$*$\#\#\#\#e&baked/bake, charged/charge\\
Xed$_{\rm PP}$&X$_{\rm V}$&$*$\#\#\#\#\#\#\#\#&$*$\#\#\#\#\#\#&directed/direct\\
Xs$_{\rm Np}$&X$_{\rm Ns}$&$*$$*$$*$$*$$*$$*$\#\#\#\#\#&$*$$*$$*$$*$$*$$*$\#\#\#\#&helmets/helmet, rabbits/rabbit\\
Xing$_{\rm GER}$&Xed$_{\rm PP}$&$*$$*$$*$$*$$*$$*$\#\#\#\#\#\#\#&$*$$*$$*$$*$$*$$*$\#\#\#\#\#\#&walking/walked, talking/talked\\
Xing$_{\rm GER}$&Xs$_{\rm V3s}$&$*$$*$$*$$*$$*$\#\#\#\#\#\#\#&$*$$*$$*$$*$$*$\#\#\#\#\#&walking/walks, talking/talks\\
Xness$_{\rm Ns}$&X$_{\rm ADJ}$&$*$$*$$*$$*$\#\#\#\#\#\#\#\#\#&$*$$*$$*$$*$$*$$*$$*$$*$\#\#\#\#\#&short/shortness\\
Xly$_{\rm ADV}$&X$_{\rm ADJ}$&$*$$*$$*$$*$$*$$*$\#\#\#\#\#\#&$*$$*$$*$$*$$*$$*$\#\#\#\#&easy/easily, quick/quickly\\
Xest$_{\rm ADJ}$&X$_{\rm ADJ}$&$*$\#\#\#\#\#\#\#&$*$\#\#\#\#&hardest/hard, shortest/short\\
Xs$_{\rm V3s}$&X$_{\rm V}$&$*$$*$$*$\#\#\#\#\#&$*$$*$$*$\#\#\#\#&jumps/jump, plays/play\\
Xer$_{\rm ADJ}$&X$_{\rm ADJ}$&$*$\#\#\#\#\#\#&$*$\#\#\#\#&harder/hard, louder/loud\\
Xless$_{\rm ADJ}$&X$_{\rm Ns}$&$*$\#\#\#\#\#\#\#\#&$*$\#\#\#\#&painless/pain, childless/child\\
Xing$_{\rm GER}$&Xy$_{\rm ADJ}$&$*$\#\#\#\#\#\#\#&$*$\#\#\#\#\#&raining/rainy, running/runny\\
Xed$_{\rm PP}$&Xs$_{\rm V3s}$&$*$$*$\#\#\#\#\#\#&$*$$*$\#\#\#\#\#&played/plays\\
Xings$_{\rm Np}$&X$_{\rm
V}$&$*$$*$$*$\#\#\#\#\#\#\#\#\#&$*$$*$$*$\#\#\#\#\#&paintings/paint\\ \hline
\end{tabular} \end{table*}

WWM then goes through the lexicon word by word and attempts to unify
each word in form and category with the left or right side of this strategy. 
If it succeeds, WWM
replaces all the segments fully specified on the side of the strategy
the word is unified with, with the segments fully specified on the other
side. For example, given the noun \emph{perception} in the corpus and strategy (\ref{eption3}), WWM will
map the word onto the right hand side of (\ref{eption3}), take out the sequence
``ception'' from the end and replace it with the sequence ``ceive'' to
produce the new word \emph{perceive.}
 The
category of the word will also be changed from singular noun to verb.
New words can thus be generated in a rather obvious fashion by taking
each word in the original lexicon and applying any strategies that can
be applied, i.e.\ whose orthographic form and part of speech can be unified with the word
at hand.
Algorithm~\ref{generate} shows the basic generation procedure; once
again the routines called \texttt{unify} and \texttt{create} which
implement the nitty-gritty details of the above description are not
given because they are more tedious than interesting, and will
certainly need to be changed in more general future versions of WWM.
Table~\ref{prince} gives some of the new words WWM creates using text from \emph{Le petit
prince} as its base lexicon.
\begin{table}[t] \caption{Words generated from \emph{Le petit prince}}
 \label{prince}
\bigskip \begin{center}
\begin{tabular}[t]{llll}
drames&Np&droitement&ADV\\ 
dress\'{e}e&PF&dr\^{o}les&AIP\\ 
dresser&INF&dr\^{o}lement&ADV\\
dressa&Vp3&dunes&Np\\ 
dressais&Vi2&durerait&Vc3\\
dresse&V3&d\'{e}cid\'{e}e&PF\\ 
dressent&V6&d\'{e}cider&INF\\ 
dressez&V5&d\'{e}cida&Vp3\\
dressait&Vi3&d\'{e}cide&V3\\
droits&AMP&d\'{e}coiff\'{e}&AM\\ 
droites&AFP&d\'{e}concentr\'{e}s&AMP
\end{tabular} \end{center} \end{table} 

The output from the algorithm is a list of words,\footnote{By
\emph{word} we mean an orthographic form together with the part of
speech.
Further work in this vein would add meanings as well.} much as in
Table~\ref{prince}, which are generated
from the input corpus using the morphological relations (strategies)
discovered.
The method described above will clearly force WWM to
create words that were already part of its original lexicon; in fact,
each and every word involved in licensing the discovery of a
morphological strategy will be duplicated by the
program.
Generated words that were \emph{not} part of WWM's original
lexicon are then added to a separate word list containing only new
words.
 If desired, this new word list can be merged with the original
lexicon for another round of discovery to formulate new strategies based on a larger
dataset.
Additionally, each of the new
words can simply be put through another cycle of word creation
by applying the same strategies as before a second time.

\section{Implementation}

This section contains some pseudocode showing several basic components
of the Whole Word Morphologizer.
Algorithm~\ref{WWMalg} shows the main procedure, which takes a
POS-tagged lexicon as input and outputs a list of all words that are
possible given the morphological relations present in the lexicon.
\begin{algorithm}[ht] \caption{\textbf{WWM}(\texttt{lexicon})} \label{WWMalg}
\begin{algorithmic}
\REQUIRE \texttt{lexicon} to be a list of POS-tagged words.
\ENSURE a list \texttt{newwords} is generated 
\FORALL{tagged words $w_i$}
\FORALL{tagged words $w_j$}
\IF{$w_i$ and $w_j$ share a beginning sequence}
\STATE $\texttt{\bf compforward}(w_i, w_j)$
\ELSIF{$w_i$ and $w_j$ share an ending sequence}
\STATE $\texttt{\bf compbackward}(w_i, w_j)$
\ENDIF
\ENDFOR
\ENDFOR
\FORALL{\texttt{comparison} structures in the list}
\IF{count(\texttt{comparison}) $>$ Threshold}
\STATE append \texttt{comparison} to the list \texttt{strategies}
\STATE \texttt{\bf generate}(\texttt{lexicon}, \texttt{strategies})
\ENDIF 
\ENDFOR
\end{algorithmic}
\end{algorithm}

The two procedures \textbf{compforward} and \textbf{compbackward} are
symmetrical, so Algorithm~\ref{compforward} shows just the first of
these.
This algorithm provides the data structure which includes the
differences and similarities between each pair of words in the
lexicon, in similar fashion to the examples in the preceding section.
In practice, only those pairs of words which are by some heuristic
sufficiently similar in the first place are compared.
Additionally, the two similarities sequences for each word pair are
actually represented as one sequence which encodes the information
found in the two sequences of the examples in the preceding; this is
just for convenience of storage and computation.
\begin{algorithm}[h] \caption{\textbf{compforward}$(w_1, w_2)$}
\label{compforward}
\begin{algorithmic}
\REQUIRE $w_1$ and $w_2$ to be (word, category) pairs.
\ENSURE a data structure \texttt{comparison} documenting the different and similar letters
between $w_1$ and $w_2$ is merged into the global list of comparisons.
\texttt{comparison} is a structure of 5 lists $w_1$dif, $w_1$cat,
$w_2$dif, $w_2$cat, sim.
\FOR{$x = 1$ to length($w_2$)}
\IF{characters $w_1(x) = w_2(x)$}
\STATE append $w_1(x)$ to list sim
\IF{list $w_1$dif does not end with `X'}
\STATE append `X' to both lists $w_1$dif and $w_2$dif
\ELSE \STATE append $w_1(x)$ to $w_1$dif,\\
append $w_2(x)$ to $w_2$dif, append `\#' to sim
\ENDIF \ENDIF
\ENDFOR
\FOR{$x = \text{length}(w_2)+1$ to length($w_1$)}
\STATE append $w_1(x)$ to $w_1$dif
\ENDFOR
\IF{dif lists and categories match a comparison already in the list \texttt{comps}}
\STATE \textbf{merge} comparisons and increment count(\texttt{comparison})
\ELSE 
\STATE append \texttt{comparison} to \texttt{comps}
\STATE count(\texttt{comparison}) $\leftarrow$ 1
\ENDIF
\end{algorithmic}
\end{algorithm}

Algorithm~\ref{generate} shows the outline of the final stage, which
generates an output list of words from the input lexicon and the
morphological strategies.
The strategy list is simply a list of all comparison structures that
occurred more frequently than some arbitrary threshold number.  
\begin{algorithm}[h] \caption{\textbf{generate}(\texttt{lexicon},
\texttt{strategies})}
\label{generate}
\begin{algorithmic}
\ENSURE a list \texttt{newwords} is generated using \texttt{lexicon} and \texttt{strategies}
\FORALL{words in \texttt{lexicon}}
\FORALL{\texttt{strategies}}
\IF{\texttt{\bf unify}(\texttt{lexicon}[x], \texttt{strategies}[x]) says the word and
strategy match with either left or right alignment}
\STATE newword $\leftarrow$ \texttt{\bf create}(\texttt{lexicon}[x],
\texttt{strategies}[x])
\IF{newword is not in the lexicon or the list \texttt{newwords}}
\STATE append newword to \texttt{newwords} list
\ENDIF
\ENDIF
\ENDFOR
\ENDFOR
\end{algorithmic}
\end{algorithm}

\section{Accomplishments and prospects}

\subsection{Initial results}

Whole Word Morphologizer has been tested on a limited basis using
English and French lexicons of approximately 3000 entries, garnered from the
POS-tagged versions of Le petit prince and Moby Dick.  The
program initially, without any post-hoc corrections, achieved between 70\% and 82\% accuracy in
generation; these figures measure the percentage of the new words beyond
the original lexicon that are possible words of the language.
The figures thus measure a kind of \emph{precision} value, in terms of
the precision/recall tradeoff, and are fair values in that they do not
include the generated words that are already in the lexicon.

A satisfactory \emph{recall} metric seems impossible to think of in
its usual sense here.
First of all, there are generally an indefinite number of
possible words in a language.
One therefore cannot give a precise set of words that we wish the system could
generate from a specific lexicon, so there seems to be no way to measure the
percentage of ``desired words'' that are in fact generated.
Even if we were to make such a list by hand from the current small
corpora to use as a gold standard (which has been suggested by a referee), it must also be remembered
that WWM discovers strategies (morphological relations) for creating new words from given ones.
It cannot be expected to discover strategies that are not evident in a
corpus.
Indeed, WWM will \emph{never} discover that, for example, `am' and `be' are related,
because according to the theory of morphology being applied these
words are only related by convention, not by morphology.
``Nonproductive morphology'' is not really morphology.

The real point is that we do not want to hold WWM's performance up against our
own ideas about morphological relations among words, since it would be
practically impossible to determine not merely a large set of possible
words that \emph{linguists} think are related to those in the corpus, but
rather a set of possible words that WWM \emph{ought to generate}
according to its theory.
This would amount to trying to beat WWM at its own game in pursuit of
a gold standard, which could only be obtained using a better
implementation of WWM's theory.
A perfect implementation of Whole Word Morphology would have perfect
recall, in view of our eventual goal of using this theory to inform us
about the morphology of a language---about what ought to be recalled. 
We are not trying to learn something that we feel is already known.

\subsection{What's learning?}

It is worth
considering the endeavor of learning morphology in terms of formal
learning theory, as presented in Osherson et al.\
\shortcite{Oshersonetal1986} or Kanazawa \shortcite{Kanazawa1998} for example.
In the classical framework, the problem of learning a language from
positive example data is approached by considering the successive guesses at
the target language that
a purported learner makes when presented with some sequentially
increasing learning sample drawn from that language.
Considering just morphology, it seems that the target language is the
set of all possible words of the natural language at hand, a possibly
infinite (or at least indefinite) set.
WWM's output is a list of generated words subsuming the corpus, which
are supposed to be all the words creatable by applying its idea of
morphology to that corpus.
It can thus be viewed as making a guess about the target language, given a
certain learning sample.
If the learning sample is increased, its guess increases in size also.
The errors in precision of course mean that at the current corpus
sizes its guesses are for the moment not even subsets of the target language.

According to one classic paradigm, a system would be held to be a
successful learner if it could be proven to home in on the target
language as the learning sample increased in size indefinitely.
This is Gold's \shortcite{Gold1967} criterion of \emph{identification in
the limit.}
In this framework, an empirical analysis cannot be used to decide the
adequacy of a learner, and we would like to deemphasize the importance
of the empirical results for this purpose.
That said, the empirical results are for now all we have to show, but eventually
we hope to produce a mathematical proof of just what WWM can learn,
and just what kinds of lexicons are learnable in Gold's sense.

To our knowledge, it has never been proven whether the total lexicon
of a natural language is identifiable in the limit from the sort of
data we provide (i.e.\ POS-tagged words), using in particular the
theory of Whole Word Morphology in a perfect fashion.
Still, it is interesting that nothing about this language
learning paradigm says anything about morphological analysis.
The current crop of true morphological learners, e.g.\
\cite{Goldsmith2001a}, endeavor to learn to analyze the morphology of
the language at hand in the manner of a linguist.
Goldsmith has even called his Linguistica system a ``linguist in a
box.''
This is perhaps an interesting and worthwhile endeavor, but it is not
one that is undertaken here.
WWM is instead attempting to learn the target language in a more
direct way from the data, without first constructing the intermediary
of a traditional morphological analysis.
We are thus not learning the linguist's notion of
morphology but rather the \emph{result} of morphology, i.e.\ the word
forms of the language together with the other information that goes
into a word.\footnote{In this theory, a word's form cannot be usefully divorced from the
other information that allows its proper use, and in our
implementation the POS tags (poor substitutes for what should be a richer database of
information) are crucial to the discovery of the strategies.}

\subsection{Post-hoc fixes and future developments}

A significant proportion of errors in generation result from the
application of competing ambiguous morphological strategies.
 For example, when
using the (French) text of \emph{Le petit prince} as its base lexicon, WWM produces
two strategies relating 2nd person verb forms to their
infinitives. 
Given the verb \emph{conjugues} `conjugate,' pres.\ 2nd sing.,
one strategy produces the correct \emph{-er} class infinitive \emph{conjuguer} while the other
creates the non-word \emph{*conjuguere}, based on the relation among
\emph{-re} verb forms like
\emph{fais/faire} `do' and \emph{vends/vendre} `sell.'
This is because of an inherent ambiguity among various word pairs
which do not fully indicate the paradigms of which they are a part.
 WWM then adds to its
lexicon, not only the correct form, but all the outputs warranted by
its grammar. 

To try to correct this problem, a form of lexical blocking has been
implemented in the current version of the program.
WWM creates every
possible word, including different strategies giving the same one, and lets
lexical lookup take precedence over productive morphology. 
The knowledge WWM possesses about its lexicon increases considerably
during the creation of morphological strategies. The program learns
not only which strategies are licensed by a given lexicon, but also
which words of its lexicon are related to one another. WWM can assign
a number to every lexical entry and give the same ``paradigm'' number to
related words.  Before adding a newly created word to its lexicon, the
program looks for an existing word with the same paradigm number and
category. For example, if WWM maps the word \emph{decoction,} which was
assigned to, say, paradigm 489 onto a strategy creating plural nouns,
it will look for a plural noun belonging to paradigm 489 in its
lexicon before it adds \emph{decoctions} to the list of new words.

Preliminary results are encouraging, with WWM reaching up to 92\%
accuracy in generation after the blocking modification.
  Obviously the program
needs to be systematically tested on multiple lexica from different
languages, but these results strongly suggest that it is possible to
model the acquisition of morphology as a component of learning to
generate language directly, rather
than to treat computational learning as the acquisition of linguistic
theory as several current approaches do, e.g.\ \cite{Goldsmith2001a}.

Although the principles of whole word morphology allow one to
contemplate versions of WWM that would work on templatic morphologies,
polysynthetic languages, and a host of other recalcitrant phenomena,
the current instantiation of the program is not so ambitious.
The comparison algorithm detailed in the previous section compares
words letter by letter, either from left to right or from right to
left.  No other possible alignments between words are considered and
WWM is in its current state only capable of grasping prefixal and
suffixal morphology.  We are currently developing a more sophisticated
sequence alignment routine which will allow the program to handle infixing,
circumfixing, and templatic morphologies of the Semitic type, as well
as word-internal changes typified by Germanic strong verb ablaut.


\bibliographystyle{acl}
\bibliography{/home/safulop/bibfiles/ling,/home/safulop/bibfiles/compling,/home/safulop/bibfiles/mathling}

\end{document}